%% file: main.tex
\documentclass[runningheads]{llncs}

\usepackage{eccv}

\usepackage{url}
\usepackage{multirow}

\usepackage{eccvabbrv}

\usepackage{graphicx}
\usepackage{booktabs}
\usepackage{algorithm}
\usepackage{algorithmic}
\usepackage{amsmath}
\usepackage[table,dvipsnames]{xcolor}
\usepackage[numbers]{natbib}
\usepackage[accsupp]{axessibility}

\usepackage{pifont}
\newcommand{\cmark}{\ding{51}}

\definecolor{darkblue}{RGB}{0,0,160}
\definecolor{royalblue}{RGB}{0,90,255}

\newcommand{\best}[1]{\textbf{\textcolor{darkblue}{#1}}}
\newcommand{\second}[1]{\textcolor{royalblue}{#1}}

\usepackage[pagebackref,breaklinks,colorlinks,citecolor=eccvblue]{hyperref}

\usepackage{orcidlink}

\begin{document}

\title{LiteMatch: Lightweight Zero-Shot Stereo Matching via Cost Volume Stabilization} 

\titlerunning{LiteMatch}

\author{Md Raqib Khan\inst{1} \and
Santosh Kumar Vipparthi\inst{2} \and
Subrahmanyam Murala\inst{1}}

\authorrunning{Khan et al.}


\institute{CVPR Lab, Trinity College Dublin, The University of Dublin, Dublin, Ireland
\and
CVPR Lab, Indian Institute of Technology Ropar, Rupnagar, Punjab, India
\email{khanmd@tcd.ie}
}

\maketitle

\begin{abstract}
Despite rapid progress in learning-based stereo matching, high accuracy is often achieved at the cost of heavy backbones and computationally intensive 3D cost volume processing, resulting in substantial memory and runtime overhead. More critically, these methods frequently struggle to generalize across domains, limiting their practical deployment. We present \textit{LiteMatch}, a lightweight stereo matching framework that achieves strong zero-shot generalization through cost volume stabilization-without expensive 3D convolutions. LiteMatch employs two complementary encoders: a Cross-View Correspondence Encoder (CVCE) to capture global cross-view interactions, and a High-Frequency Encoder (HFE) that enhances fine structural details via FFT-based frequency cues. To stabilize the cost volume, we introduce the \textit{Cost Volume Consistency Loss (CVC-Loss)}, a voxel-wise binary cross-entropy objective applied to softmax-normalized cost distributions. By encouraging sharp and unimodal disparity probabilities, CVC-Loss promotes stable cost distributions and enables rapid convergence. A lightweight refinement module further produces sharp full-resolution disparities with low-iteration updates, avoiding heavy recurrent refinement. With a flexible design ranging from 3.36M to 9.58M parameters, LiteMatch achieves exceptional zero-shot generalization, delivering competitive EPE and D1 performance across Scene Flow, KITTI, Middlebury, ETH3D, and DrivingStereo. Our results establish that lightweight architectures can indeed generalize across domains without sacrificing accuracy. \href{https://mdraqibkhan.github.io/Litematch}{\textcolor{blue}{Code}}
\end{abstract}

\section{Introduction}
\begin{figure}[!ht]
    \centering
\includegraphics[width=1\textwidth]{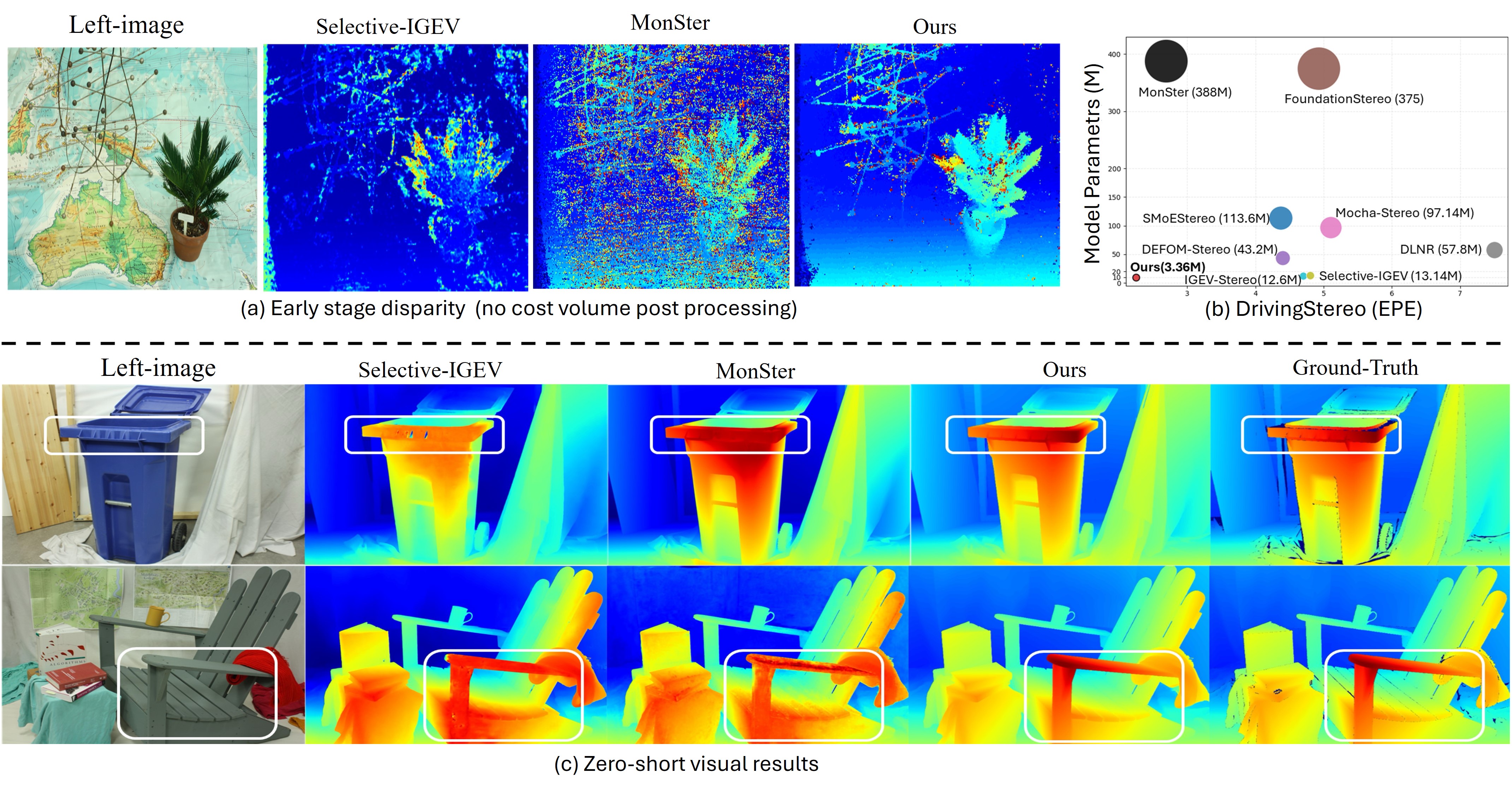}
   \caption{ Zero‑shot cross‑domain generalization. (a) \& (c) LiteMatch produces cleaner disparity maps than state‑of‑the‑art methods on challenging real‑world scenes, (b) despite having significantly fewer parameters. This demonstrates superior efficiency and robustness to domain shift.}
\label{fig:cost_volume_consistency}
\vspace{-3mm}
\end{figure}
\vspace{-2mm}
\label{sec:intro}
Stereo depth estimation is a cornerstone of 3D perception, with critical applications in autonomous driving, robotics, and augmented reality (AR)~\cite{chen2025feature,zbontar2015computing}. The task demands pixel-accurate disparity maps, particularly at object boundaries, while simultaneously meeting the strict real-time constraints of embedded systems~\cite{scharstein2002taxonomy}. 

The advent of deep learning has fundamentally transformed this field. Early models like PSMNet~\cite{chang2018pyramid} and GA-Net~\cite{zhang2019ga} established a powerful paradigm: building a 4D cost volume followed by regularization using 3D convolutional neural networks (CNNs). While effective at encoding geometry, this approach incurs prohibitive computational and memory costs due to the cubic complexity of 3D convolutions. This limitation has motivated the exploration of more efficient paradigms. RAFT-Stereo~\cite{lipson2021raft} pioneered a different strategy, replacing volumetric processing with iterative refinement driven by a high-resolution correlation pyramid. This achieved scalability but traded one bottleneck for another: its lack of a strong geometric prior led to slow convergence and persistent errors in low-texture regions~\cite{xu2023iterative}. Subsequent works like IGEV-Stereo~\cite{xu2023iterative} and Selective-IGEV~\cite{wang2024selective} addressed this by hybridizing the approaches, using lightweight 3D cost volumes to generate robust geometry encodings for initializing iterative modules. Nevertheless, this hybrid paradigm still relies on 3D convolutions, preserving significant computational overhead.

Alternatively, a second paradigm leverages massive foundation models and monocular depth priors~\cite{cheng2025monster,wen2025foundationstereo, jiang2025defomstereo}. While achieving impressive performance, these methods require hundreds of millions of parameters and exhibit slow inference speeds that preclude real-time deployment.

We identify a critical, underlying flaw common to these diverse state-of-the-art approaches: \textit{a fundamental instability in the early feature matching process}. As visualized in Fig.~\ref{fig:cost_volume_consistency} (a), even models with large, pre-trained backbones produce noisy and ambiguous cost volumes. This initial instability creates a dependency cycle: noisy cost volumes necessitate either heavy 3D regularization or dozens of iterative refinement steps to recover accurate disparities. Consequently, the field has predominantly focused on designing increasingly powerful corrective modules (large 3D CNNs, recurrent GRUs), while the root cause of the noise remains unaddressed. This post-processing overhead constitutes the primary bottleneck for efficiency and real-time deployment.

This analysis leads to a fundamental question: \textit{Can we achieve robust, real-time stereo matching by stabilizing the feature representation and cost volume at the source}, thereby eliminating dependency on 3D convolutions, large pre-trained networks, and external monocular priors?

\begin{figure*}[t!]
\centering
\includegraphics[width=0.95\textwidth]{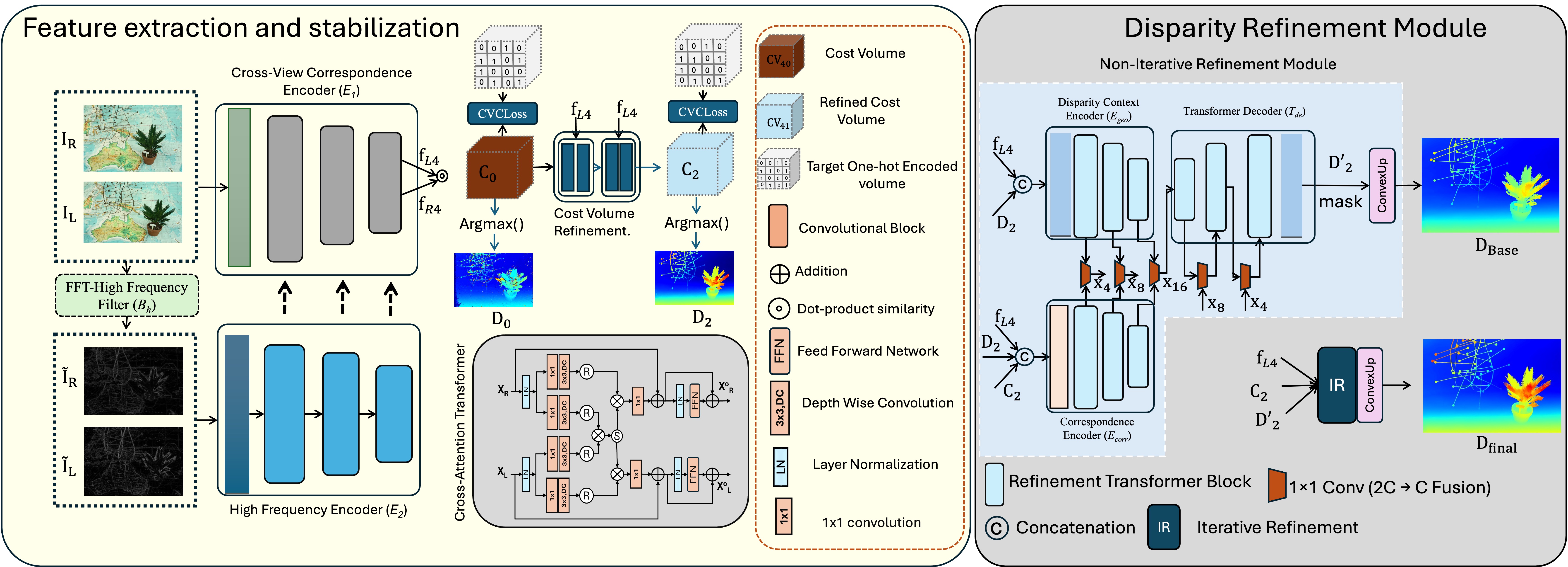}
\caption{Overview of LiteMatch. Our framework employs a progressive two-stage scheme: (1) coarse training jointly optimizes the feature extractor and cost volume using our CVC-Loss to establish stable cost volume representations; (2) refinement training freezes Stage 1 components and trains only the transformer-based refinement module with L1 loss to recover fine details. The default model performs efficient single-pass inference, with iterative refinement (IR) extension for comparison.}
\label{fig:maindiagram}
\end{figure*}
To address this challenge, we introduce LiteMatch, a novel stereo matching framework designed to prevent cost volume instability at its source. Rather than developing more powerful regularizers, LiteMatch ensures feature stability and cost volume sharpness from the beginning. This principled approach eliminates the need for 3D convolutions or large backbones, establishing an improved balance between performance and complexity.
Our main contributions are summarized as follows:
\begin{itemize}

    \item We propose \textbf{LiteMatch}, a lightweight zero-shot stereo matching framework that operates without large backbone networks or monocular depth prior guidance.
    
    \item We introduce a frequency-aware stereo representation that integrates prominent correspondence features with high-frequency structural cues, improving disparity estimation in structurally ambiguous regions.
    
    \item We propose CVC-Loss, a distribution-level supervision strategy that stabilizes cost volume representations by enforcing inter-bin contrast across disparity hypotheses, producing sharper and more discriminative initial predictions.
    
    \item We design a lightweight disparity refinement framework that achieves strong single-pass performance and further attains state-of-the-art accuracy when extended with iterative refinement.

\end{itemize}
We evaluate LiteMatch extensively on five benchmark datasets (KITTI 2012/2015, Middlebury, ETH3D, and DrivingStereo). Results demonstrate that LiteMatch achieves superior performance and cross-domain generalization despite its lightweight design,  with comprehensive ablation study validating the contribution of each core component of our proposed approach.

\section{Related Work}
\label{sec:related}
\vspace{-3mm}

\noindent\textbf{Cost-Volume Regularization and Distribution Modeling.}  
Cost volume construction and regularization remain central to modern stereo matching. Early deep stereo methods relied heavily on stacked 3D CNNs to suppress noisy correlations and enforce spatial consistency, achieving strong accuracy at the expense of computational cost. To improve efficiency, several works have explored alternative cost-volume processing strategies. CFNet~\cite{shen2021cfnet} leverages multi-resolution cost volumes to better handle large disparity ranges. ACFNet~\cite{zhang2020adaptive} promotes unimodal disparity distributions through adaptive filtering mechanisms. Bi3D~\cite{badki2020bi3d} reformulates stereo estimation as a sequence of binary disparity threshold classification problems for real-time inference. More recently, ADL~\cite{xu2024adaptive} models disparity boundary uncertainty using Laplacian mixture distributions to better capture multi-modal ambiguities. These approaches reduce computational overhead or improve uncertainty modeling, yet most methods still depend on either volumetric processing, iterative refinement, or additional mechanisms to mitigate ambiguous cost responses. Designing efficient strategies that maintain stable and well-structured disparity distributions without heavy regularization remains an active research direction.

\noindent\textbf{Feature Extraction Architectures.}  
Feature representation quality strongly influences the reliability of stereo matching. Recent works increasingly adopt large pretrained backbones or monocular priors to enhance cross-domain robustness~\cite{cheng2025monster, bartolomei2025stereo, wen2025foundationstereo, yao2025diving}. While effective, such approaches typically incur substantial computational and memory overhead. Lightweight alternatives aim to improve efficiency by simplifying backbone design or increasing spatial downsampling (e.g., DLNR~\cite{zhao2023high}), which may reduce fine-grained detail preservation and amplify ambiguity in low-texture or repetitive regions. Balancing efficiency and feature fidelity remains a key challenge in lightweight stereo architectures, particularly when striving for robust generalization across diverse domains.

\noindent\textbf{Iterative Refinement Paradigms.}  
Iterative update frameworks have become dominant in high-accuracy stereo systems. RAFT-Stereo~\cite{lipson2021raft} introduced recurrent disparity refinement using correlation volumes and gated update operators. Subsequent works enhance this paradigm through normalization techniques (DLNR~\cite{zhao2023high}), motif-based attention mechanisms (MoCha-Stereo~\cite{chen2024mocha}), frequency-adaptive fusion strategies (Selective-IGEV~\cite{wang2024selective}), and integration of monocular priors during refinement (MonSter~\cite{cheng2025monster}). These approaches achieve state-of-the-art performance but often require multiple refinement iterations, increasing inference latency. An alternative design is An alternative design is HITNet~\cite{tankovich2021hitnet}, which avoids explicit 3D cost volumes by employing hierarchical tile-based refinement with planar patch propagation and geometric warping. This structured geometric reasoning enables real-time performance with a compact parameter count. 

Across both volumetric and geometric paradigms, a fundamental trade-off persists between computational efficiency and cross-domain generalization. Architectures optimized for lightweight inference often simplify feature extraction or refinement, potentially weakening robustness in ambiguous or unseen environments. Conversely, models designed for strong generalization frequently rely on large backbones, iterative updates, or heavy regularization, increasing computational cost. Balancing these competing objectives remains a central challenge in modern stereo matching.

\section{Methodology}
\label{sec:methodology}
\vspace{-3mm}
An overview of our framework is illustrated in Figure~\ref{fig:maindiagram}. We propose LiteMatch, a lightweight and efficient stereo matching framework that achieves high performance without relying on large backbone or depth prior. LiteMatch follows a two-stage design comprising three core components: (1) a dual- branch encoder for stable and geometry-aware feature extraction, (2) the CVC-Loss to stabilize the cost volume during early stages of training, and (3) disparity refinement module is non-iterative for faster inference, but can be optionally extended with an iterative process for fair SOTA comparison. 
Importantly, our iterative refinement achieves comparable performance with fewer iterations than existing methods.

\noindent\textbf{Pipeline Overview:}
Given a rectified stereo pair ${I_L, I_R} \in \mathbb{R}^{H \times W \times 3}$, LiteMatch estimates the final disparity map $D_{\text{final}} \in \mathbb{R}^{H \times W}$ through a two-stage training scheme. In Stage-1, the initial cost volume $C_0$ is computed using correlation between features $f_{L4}$ and $f_{R4}$ from the dual-encoder, and $C_2$ is calculated through cost volume refinement (Sec.~\ref{subsec:cvc_loss}).
 In Stage 2, the Stage-1 parameters are frozen to ensure stability, and a disparity refinement module is trained to take refined $C_{2}$ and left feature $f_{L4}$ as inputs, generating a high-resolution disparity map $D_{\text{base}}$ and its iterative refinement $D_{\text{final}}$.
\vspace{-2mm}
\subsection{Feature Extraction (Dual Encoders)}
\label{subsec:dual_encoder}
\vspace{-2mm}
Feature quality critically affects stereo matching performance, yet existing backbones face two core limitations. On one hand, convolutional networks~\citep{wang2024selective} provide strong local priors but are limited by small receptive fields, causing oversmoothing in low-texture regions. On the other hand, transformer-based encoders~\citep{zhao2023high,khan2025phaseformer, khan2024spectroformer, khan2023underwater} capture global context but lose spatial precision due to global attention. To overcome these limitations, LiteMatch employs a dual encoder that brings together two complementary inductive biases, a Cross-view Correspondence Encoder ($E_1$) for long-range correspondence modeling, and a High-Frequency Encoder ($E_2$) for edge and texture preservation \cite{wang2022domain}. Fusing these representations yields stereo features that are both globally coherent and spatially sharp, improving disparity prediction under diverse conditions.

\subsubsection{Cross-view Correspondence Encoder ($E_1$):}
\label{subsec:cross_attention}
\vspace{-0.5mm}

It uses a multi-head cross-attention (CA) mechanism to establish dense correspondence between the left ($\mathbf{X}_L$) and right ($\mathbf{X}_R$) features, where $\mathbf{X}_{L/R} \in \mathbb{R}^{B \times C \times H \times W}$. We generate query ($\mathbf{Q}$), key ($\mathbf{K}$), and value ($\mathbf{V}$) tensors from the input features:
{\footnotesize
\begin{align}
\mathbf{Q} &= \varphi_3(\psi_1(\mathrm{LN}(\mathbf{X}_L))), \quad \mathbf{K} = \varphi_3(\psi_1(\mathrm{LN}(\mathbf{X}_R)))
\\
&      \mathbf{V}_{L/R} = \varphi_3(\psi_1(\mathrm{LN}(\mathbf{X}_{L/R})))
\end{align}
}

where $\psi_1$ and $\varphi_3$ are $1\times1$ and depth-wise $3\times3$ convolutions, respectively. The left-to-right attention map $\mathcal{A}$ is computed as:
\begin{equation}
\mathcal{A} = \mathrm{Softmax}\!\left(\frac{\mathbf{Q}\cdot \mathbf{K}^\top}{\sqrt{\alpha}}\right),
\end{equation}
where $\alpha$ is is a learnable scale controlling the magnitude of $\mathbf{Q} \cdot \mathbf{K}^\top$ prior to softmax. We calculate both views using bidirectional attention, leveraging the transpose $\mathcal{A}^\top$ for right-to-left alignment:
\begin{align}
\mathbf{X}_{L/R}^{\mathrm{out}} &= \psi_1(\mathcal{A}\mathbf{V}_{L/R}) + \mathbf{X}_{L/R}
\end{align}
Residual connections and FFNs~\citep{Zamir2021Restormer} are applied to $\mathbf{X}_{L/R}^{\mathrm{out}}$ to stabilize training and enhance correspondence awareness.

\subsubsection{High-Frequency Encoder ($E_2$)}
\label{subsec:high_frequency}
\vspace{-1mm}

The High-Frequency Encoder ($E_2$) is a convolutional network designed to preserve high-frequency features (edges and texture). It processes inputs that have been pre-filtered to isolate high-frequency content, $\tilde{I}_L, \tilde{I}_R = \mathcal{H}(I_L, I_R)$, where $\mathcal{H}(\cdot)$ uses a Fourier-based high-pass filter~\citep{tan2024frequency}. The filtering operation is defined as:
\begin{align}
\tilde{I} &= \mathcal{F}^{-1}\!\big(B_h(\mathcal{F}(I))\big), \\
B_h(f_{u,v}) &=
\begin{cases}
0, & |u| < W/4,\, |v| < H/4,\\
f_{u,v}, & \text{otherwise}.
\end{cases}
\end{align}
Here, $\mathcal{F}$ and $\mathcal{F}^{-1}$ denote the FFT and its inverse, respectively, and $B_h$ eliminates the low-frequency components in the central region of the frequency domain. The output features of this encoder are denoted $\mathbf{H}_{L/R}$.
\subsubsection{Fusion of Complementary Features}
\label{subsec:fusion_mechanism}
To maintain spatial sharpness, we fuse the high-frequency convolutional features ($\mathbf{H}$) with the globally coherent transformer outputs ($\mathbf{X}^{\mathrm{out}}$) using a learnable, adaptive gating mechanism. The gate $\mathbf{G}$ determines the optimal contribution of the high-frequency stream:
{\footnotesize
\begin{align}
\mathbf{G}_{L/R} &= \sigma\big(\psi_g(\mathrm{Cat}(\mathbf{X}_{L/R}^{\mathrm{out}}, \mathbf{H}_{L/R}))\big)
\end{align}
}

where $\psi_g$ is a $1\times1$ convolution and $\sigma$ is the sigmoid activation function. The final fused features ($\mathbf{F}$) are calculated as: \begin{equation}
\mathbf{F}_{L/R} = \mathbf{X}_{L/R}^{\mathrm{out}} + \lambda \mathbf{G}_{L/R} \odot \mathbf{H}_{L/R}, 
\end{equation}

The scaling factor $\lambda$ is initialized to zero and learned during training. This adaptive gating mechanism effectively balances global coherence and edge fidelity with negligible computational overhead.

\subsection{Cost Volume Stabilization}
\label{subsec:cvc_loss}
\vspace{-3mm}

A major bottleneck in stereo matching is heavy 3D cost volume regularization. LiteMatch eliminates this via two components: a Cost Volume Consistency Loss (CVC-Loss) that directly regularizes the disparity distribution, and a lightweight 2D refinement module that filters noisy correspondences.

\noindent {\textbf{Cost Volume Refinement:}}
The refinement from $C_0$ to $C_2$ is a lightweight 2D feature-guided process defined as:
\begin{equation}
    C_{(n)} = C_{(n-1)} + \phi_i \bigl( [C_{(n-1)}, f_{L4}] \bigr), \quad i \in \{1, \dots, 5\}, \quad n \in \{1, 2\}
\end{equation}
where $[\cdot]$ is channel-wise concatenation and $\phi_i$ represents a $3\times3$ convolution followed by LeakyReLU.
\begin{figure}[!ht]
    \centering
    \includegraphics[width=0.9\textwidth]{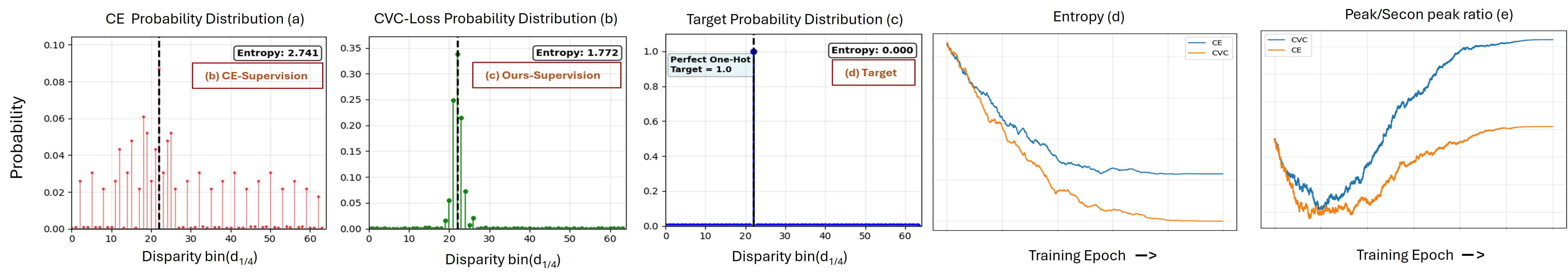}
    \vspace{-3mm}
    \caption{Cost volume distributions and entropy for a selected pixel. CE (a) shows higher entropy (2.74) than CVC (b, 1.77), with the target one-hot (c) at 0.0. Training dynamics: (d) CVC reduces uncertainty faster, and (e) maintains higher confidence via peak-to-second-peak ratios.}
    \vspace{-8mm}
    \label{fig:entropy_distribution}
\end{figure}

\noindent {\textbf{CVC-Loss Formulation:}}
Given cost volume $C\!\in\!\mathbb{R}^{H\times W\times D}$ and ground-truth $D_{\mathrm{gt}}$, we obtain probabilities $P_{i,j,d}\!=\!\mathrm{Softmax}_d(C_{i,j,d})$. Let $d^*(i,j)$ be the quantized ground-truth bin and $Y$ the corresponding one-hot volume. CVC-Loss applies voxel-wise binary cross-entropy:

\vspace{-2mm}
{\small
\[
\mathcal{L}_{\mathrm{CVC}}
=-\frac{1}{\sum M\!\cdot\! D}\!\sum_{i,j}M_{i,j}\!\sum_{d}\!
\Big( Y\log\tilde{P}+(1\!-\!Y)\log(1\!-\!\tilde{P})\Big),
\]
}
\vspace{-1mm}
\noindent where $\tilde{P}$ is a stabilized probability and $M$ masks invalid pixels.

For a pixel with ground-truth bin $d^*$, CVC-Loss decomposes as:
\[
\mathcal{L}_{\mathrm{CVC}} = \underbrace{-\log p_{d^*}}_{\text{standard CE}} \;+\!\! \sum_{d\neq d^*} \underbrace{-\log(1-p_d)}_{\text{explicit suppression}}.
\]
Thus, CVC-Loss \emph{simultaneously} pushes up the correct probability \emph{and} pulls down every incorrect hypothesis. The gradient for an incorrect bin $k\neq d^*$ is $\frac{\partial}{\partial p_k}(-\log(1-p_k)) = \frac{1}{1-p_k}$, which grows rapidly as $p_k\!\to\!1$, creating an adaptive suppression mechanism. Using $-\log(1-x)\ge x$ for $x\in[0,1)$, we obtain the lower bound:
\[
\mathcal{L}_{\mathrm{CVC}} \ge -\log p_{d^*} + (1-p_{d^*}).
\]
This shows CVC-Loss \emph{jointly} maximizes the correct probability while minimizing total mass on incorrect bins, producing a provably sharper cost volume than CE alone.

\noindent{\textbf{Empirical Behavior and Training Stability:}}
While categorical cross-entropy (CE) is commonly used for disparity classification, it provides weak supervision for incorrect bins, leading to noisy, multi-modal cost volumes in early training. In contrast, CVC-Loss supplies dense negative supervision across \emph{all} disparity bins via $(1-Y_{i,j,d})\log(1-\tilde{P}_{i,j,d})$, actively suppressing incorrect hypotheses. This yields three measurable benefits (Figs.~\ref{fig:entropy_distribution}a and \ref{fig:entropy_distribution}b):
(i) {35.4\% lower entropy} (1.77 vs. 2.74 for CE);
(ii) faster uncertainty reduction during training (Fig.~\ref{fig:entropy_distribution}d); and
(iii) {higher confidence} via improved peak-to-second-peak ratios (Fig.~\ref{fig:entropy_distribution}e).

\subsection{Disparity Refinement Module}
\label{sec:refinement}
\vspace{-2mm}
Our refinement strategy emphasizes both high speed and accuracy without requiring mandatory iterative processing. We achieve this through a base non-iterative model, which can be optionally augmented with an IR head for SOTA comparisons

\noindent\textbf{Non-Iterative Disparity Refinement (Base Model):}
\label{subsubsec:non_iterative}
Our base model employs a highly efficient, non-iterative refinement module that enhances disparity in a single forward pass. The input to the module consists of the disparity $D_{2}$, left-view image features $f_{L4}$, and the refined cost volume $C_{2}$. The refinement network follows a transformer-based dual-encoder and single-decoder design, leveraging complementary input streams. The Disparity-Context Encoder ($E_{\text{geo}}$) aggregates local image features $f_{L4}$ with the coarse disparity $D_2$, preserving fine textures and spatial structures around the current prediction. The Correspondence Encoder ($E_{\text{corr}}$), inspired by RAFT’s motion encoder~\citep{teed2020raft}, fuses the refined cost volume $C_2$ with $D_2$ and $f_{L4}$ to capture global correspondence cues within a transformer framework.

The multi-scale outputs from both encoders, $E_{\text{geo}, i}$ and $E_{\text{corr}, i}$, are fused by a convolutional operator $\Psi$ and passed through a hierarchical transformer decoder $T_{\text{de}}$. This decoder generates an intermediate refined disparity $D'_{2}$ and a learned convex upsampling mask (M):
\begin{align}
\text{M}, D'_{2} &= T_{\text{de}}\Psi(\{[E_{\text{geo}, i}, E_{\text{corr}, i}]\}), \quad i \in \{4, 8, 16\}
\\
D_{\text{base}} &= \mathcal{U}(D'_{2}, \text{M}),
\end{align}
where, $\mathcal{U}$ represents the convex upsampling operation. This design generates the full-resolution output $D_{\text{base}}$ in a single, efficient pass, enabling real-time disparity estimation.

\noindent\textbf{Iterative Refinement (IR) Extension:}\label{subsubsec:iterative}
For a comprehensive comparison with state-of-the-art iterative methods, we provide an \textit{iterative refinement} (IR) head. This head can be appended to the base model to enable multi-step refinement, following a GRU-based update mechanism similar to that of RAFT-Stereo~\cite{lipson2021raft}. The IR head selectively refines uncertain regions using the intermediate disparity, the left feature map, and the cost volume at one-fourth resolution. The refinement process is initialized from the coarse prediction ($\hat{D}_0 = D'_{2}$) and iteratively updated as:
\begin{align}
\hat{D}_{t+1} &= \mathrm{IR}(\hat{D}_t, f_{L_4}, C_{2}), \quad t \in \{0, 1, \dots, n-1\}
\end{align}
Unlike existing SOTA iterative frameworks, our IR converges significantly faster, requiring fewer refinement steps to achieve comparable performance. This dual-mode design provides flexibility to evaluate LiteMatch in both lightweight and high-performance regimes, effectively narrowing the gap with computationally heavier iterative methods.
\vspace{-1mm}

\subsection{Total Loss}
\vspace{-2mm}
Our two-stage training scheme uses distinct objectives for each stage.

\noindent\textbf{Stage 1} applies deep supervision to the cost volumes:
{\footnotesize
\begin{equation}
L_{\text{stage1}} = \sum_{i=0,2} \big(\mathit{Smooth}_{L1}(D_{i} - D_{\text{gt4}}) + L_{\text{CVC}}(C_{i}, D_{\text{gt4}})\big)
\end{equation}
}
where, $L_{\text{CVC}}$ denotes the CVC-Loss (see Sec.~\ref{subsec:cvc_loss}), applied to both the initial ($i=0$) and refined ($i=2$) cost volumes $C_i$. We also apply the smooth $L_1$ loss to the corresponding disparities $D_0$ and $D_2$, respectively, where $D_{gt4}$ is the ground truth at one-fourth resolution.

\noindent\textbf{Stage 2} freezes the feature and cost volume modules, and trains the refinement module with a multi-stage regression loss:
{\small
\begin{equation}
L_{\text{stage2}} = \mathit{Smooth}_{L1}(D_{\text{base}} - D_{\text{gt}}) + \sum_{j=1}^{n} \gamma^{n-j} \| D_j - D_{\text{gt}} \|_1,
\end{equation}
}
Here, $D_{\text{base}}$ denotes the base final output, $D_j$ are the intermediate predictions, $\gamma = 0.9$, $D_{\text{gt}}$ is the ground-truth disparity, and $n$ indicates the number of iterations. Each loss is used exclusively during its respective training phase. This hybrid supervision ensures stable, geometry-aware learning in Stage 1 and precise, high-resolution disparity in Stage 2, while maintaining real-time performance.

\begin{table}[t]
\centering
\tiny
\setlength{\tabcolsep}{1pt}
\vspace{-1mm}
\caption{Zero-shot D1-all (\%) evaluation on DrivingStereo and in-domain evaluation on Scene Flow. All models are trained exclusively on Scene Flow. MP and BB denote Monocular Prior and Backbone, respectively.}
\vspace{-1mm}

\begin{tabular}{l|c c c c c||l|c c c c c}
\toprule
\multicolumn{6}{c||}{\textbf{DrivingStereo (Zero-Shot)}} & \multicolumn{6}{c}{\textbf{Scene Flow (In-Domain)}} \\
\midrule
\textbf{Method} & Sunny$\downarrow$ & Cloudy$\downarrow$ & Rainy$\downarrow$ & Foggy$\downarrow$ & Avg.$\downarrow$ &
\textbf{Method} & BB/MP & Pub. & EPE$\downarrow$ & D1 (\%)$\downarrow$ & Par(M)$\downarrow$ \\
\midrule
\rowcolor{gray!18}
\multicolumn{6}{c||}{Without Monocular Depth Prior} &
\multicolumn{6}{c}{Without Monocular Depth Prior} \\
CFNet & 5.4 & 5.8 & 12.0 & 6.0 & 7.3 &
RAFT-Stereo & -- & 3DV'21  & 0.53  & 6.08  & 11.1 \\
PCWNet & 5.6 & 5.9 & 11.8 & 6.2 & 7.4 &
DLNR & -- & CVPR'23 & 0.48  & 5.06  & 57.4 \\
DLNR & 27.1 & 28.3 & 34.5 & 29.0 & 29.8 &
IGEV-Stereo & BB & CVPR'23 & 0.47  & 5.21  & 12.6 \\
IGEV-Stereo & 5.3 & 6.3 & 21.6 & 8.0 & 10.3 &
Selective-IGEV & BB & CVPR'24 & 0.44  & 4.98  & 13.1 \\
Selective-IGEV & 7.0 & 8.0 & 18.4 & 12.9 & 11.1 &
Mocha-Stereo & BB & CVPR'24 & \second{0.41} & 2.35 & 97.1 \\
Mocha-Stereo & 12.8 & 27.4 & 24.6 & 22.8 & 21.9 &
\best{LiteMatch(Base)} & -- & - & 0.53 & \second{2.29} & \best{3.36} \\
\best{LiteMatch} &\best{2.01} & \best{2.05} & \best{2.44} & \best{1.51} & \best{2.00} &
\best{LiteMatch} & -- & - & \best{0.39} & \best{1.92} & \best{9.58} \\
\midrule
\rowcolor{gray!18}
\multicolumn{6}{c||}{With Monocular Depth Prior} &
\multicolumn{6}{c}{With Monocular Depth Prior} \\
SMoEStereo & 3.3 & 3.0 & 6.0 & 4.7 & 4.3 &
DEFOM-Stereo & MP & CVPR'25 & 0.42 & 5.57 & 47.3 \\
DEFOM-Stereo &  \second{2.05} & 2.50 & \textbf{0.95} & 2.68 &  \second{2.05} &
SMoEStereo & MP & ICCV'25 & 0.50 & 5.55 & 113 \\
FoundationStereo & 3.22 & 2.81 & 11.20 &  \second{2.50} & 4.93 &
FoundationStereo & MP & CVPR'25 & \best{0.34} & - & 375 \\
MonSter & 2.60 & \second{2.12} & 3.08 & 2.94 & 2.69 &
MonSter & MP & CVPR'25 & \second{0.37} & \second{2.02} & 388 \\
\best{LiteMatch} &\best{2.01} & \best{2.05} & \second{2.44} & \textbf{1.51} & \best{2.00} &
\best{LiteMatch} & -- & - & 0.39 & \best{1.92} & \best{9.58} \\
\bottomrule
\end{tabular}
\label{tab:driving_sf_merged_single}
\vspace{-4mm}
\end{table}

\begin{table}[t]
\centering
\tiny
\setlength{\tabcolsep}{1pt}
\caption{Zero-shot performance on KITTI 2012 and KITTI 2015 (left, \textbf{Single-Pass, Iter~1}), and in-domain fine-tuned leaderboard results on ETH3D (right). All models are trained on Scene Flow for zero-shot evaluation.}
\vspace{-3mm}
\begin{tabular}{l|c c c c c||l|c c c c c c}
\toprule
\multicolumn{6}{c||}{\textbf{Zero-Shot (Scene Flow → KITTI)}} &
\multicolumn{6}{c}{\textbf{Fine-Tuned (ETH3D Leaderboard)}} \\
\midrule
\textbf{Method} & \multicolumn{2}{c}{KITTI-15} & \multicolumn{2}{c}{KITTI-12} & FPS &
\textbf{Method} & \multicolumn{3}{c}{ETH3D-Noc} & \multicolumn{3}{c}{ETH3D-All} \\
\cmidrule(lr){2-3} \cmidrule(lr){4-5} \cmidrule(lr){8-10} \cmidrule(lr){11-13}
 & EPE & Bad3.0 & EPE & Bad3.0 &  & & Bad1.0 & AvgErr & RMS & Bad1.0 & AvgErr & RMS \\
\midrule
PSMNet & 8.50 & 16.3 & 8.0 & 15.1 & 3.6 & HITNet & 2.79 & 0.20 & 0.46 & 3.11 & 0.22 & 0.55 \\
CFNet  & 4.70 & - & 5.80 & - & - & RAFT-Stereo & 2.44 & 0.18 & 0.36 & 2.60 & 0.19 & 0.42 \\
PCWNet & 4.20  & - & 5.60  & - & 2.8 & IGEV-Stereo & 1.12 & 0.14 & 0.34 & 1.51 & 0.20 & 0.86 \\
RAFT-Stereo & 3.08 & 13.5 & 2.74 & 14.1 & 11.6 & IGEV++ & 1.14 & 0.13 & 0.34 & 1.58 & 0.19 & 0.74 \\
IGEV-Stereo & 1.35 & 6.67 & 1.16 & 6.06 & 16.1 & Selective-IGEV & 1.23 & {0.12} & 0.29 & 1.56 & 0.15 & 0.57 \\
DLNR & 15.2 & 48.4 & 7.89 & 30.5 & 13.0 & DEFOM-Stereo & {0.70} & \second{0.11} & {0.22} & {0.78} & \best{0.11} & \best{0.26} \\
Selective-IGEV & 1.34 & 6.68 & 1.23 & 6.78 & 14.5 & SMoE-Stereo & 0.95 & 0.13 & 0.29 & 1.13 & {0.14} & {0.36} \\
Mocha-Stereo & 1.42 & 6.80 & 1.32 & 6.95 & 9.13 & MonSter & \second{0.44} & \best{0.10} & \second{0.20} & \second{0.70} & {0.13} & 0.47 \\
\textbf{LiteMatch (Base)} & \best{1.21} & \best{5.40} & \best{1.11} & \best{5.12} & \best{22.2} & \textbf{LiteMatch} & \best{0.35} & \second{0.11} & \best{0.19} & \best{0.52} &\second{0.12} & \second{0.31} \\
\bottomrule
\end{tabular}

\label{tab:kitti_eth3d_merged_single_best}
\vspace{-1mm}
\end{table}

\begin{table*}[!htb]
\vspace{-2mm}
\caption{
Zero-shot generalization performance on KITTI-2012, KITTI-2015, Middlebury-F, and ETH3D datasets under \textbf{default iterative settings}, trained on Scene Flow.
Inference time is measured on KITTI-2015 using an RTX A6000 at $384 \times 1248$ resolution.
\best{\textbf{Blue bold}} indicates the best result and \second{blue} indicates the second-best.
}
\vspace{-3mm}
\label{tab:comparative}
\centering
\scriptsize
\setlength{\tabcolsep}{1.5pt} 
\resizebox{\textwidth}{!}{%
\begin{tabular}{l c c c c c c c c c c c c}
\toprule
\textbf{Method} 
& \textbf{Pub.} 
& \textbf{MP/BB}  
& \textbf{K12} 
& \textbf{K15} 
& \multicolumn{2}{c}{\textbf{Middlebury-F}} 
& \multicolumn{2}{c}{\textbf{ETH3D}}
& \multicolumn{4}{c}{\textbf{Model Complexity}} \\
\cmidrule(lr){6-7} \cmidrule(lr){8-9} \cmidrule(lr){10-13}
& & & Bad3.0 & Bad3.0 & EPE & Bad2.0 & EPE & Bad1.0 
& Params & GFLOPs & Mem & Time \\
\midrule

\rowcolor{gray!15}
\multicolumn{13}{c}{\textbf{Without Monocular Depth Prior}} \\
\midrule
PSMNet~\cite{chang2018pyramid} & CVPR-18 & - & 15.1 & 16.3 & 40.51 & 57.93 & - & 23.8 & \second{5.22} & 939.6 & 7.98 & 332 \\
GWCNet~\cite{guo2019group} & CVPR-19 & - & 12.0 & 11.7 & - & 32.2 & - & 14.1 & 6.43 & 901.9 & 9.38 & 421 \\
RAFT-Stereo~\cite{lipson2021raft} & 3DV-21 & - & 5.90 & 5.86 & 5.59 & 19.5 & 0.36 & \second{3.30} & 11.11 & 525.9 & 7.08 & 353 \\
IGEV-Stereo~\cite{xu2023iterative} & CVPR-23 & BB & 5.19 & 6.06 & \second{4.29} & \best{15.1} & 0.33 & 4.00 & 12.60 & 330.6 & 6.06 & 410 \\
DLNR~\cite{zhao2023high} & CVPR-23 & - & 10.6 & 16.0 & 6.02 & 18.3 & 9.80 & 23.0 & 57.38 & 482.7 & 6.25 & 452 \\
Selective-IGEV~\cite{wang2024selective} & CVPR-24 & BB & 5.64 & 6.05 & 5.08 & 17.6 & 0.33 & 6.10 & 13.14 & 338.3 & 6.32 & 432 \\
Mocha-Stereo~\cite{chen2024mocha} & CVPR-24 & BB & \second{4.83} & 6.01 & 6.08 & \second{17.1} & \second{0.28} & 4.02 & 97.14 & 739.6 & 9.79 & 420 \\
\textbf{LiteMatch (Base)} & - & - & 5.12 & \second{5.40} & 4.60 & 18.1 & 0.33 & 5.01 & \best{3.36} & \best{79.8} & \best{1.17} & \best{45} \\
\textbf{LiteMatch} & - & - & \best{4.20} & \best{4.09} & \best{4.11} & \best{15.1} & \best{0.21} & \best{2.05} & 9.58 & \second{179.0} & \second{1.41} & \second{220} \\
\midrule
\rowcolor{gray!15}
\multicolumn{13}{c}{\textbf{With Monocular Depth Prior}} \\
\midrule
SMoEStereo~\cite{wang2025learning} & ICCV-25 & MP & 4.22 & 4.93 & 4.22 & 18.9 & \second{0.23} & 2.10 & 113.6 & 630 & 6.70 & 449 \\
DEFOM-Stereo~\cite{jiang2025defomstereo} & CVPR-25 & MP & 4.29 & 5.29 & 6.20 & 16.6 & 0.27 & 2.60 & 374.4 & 2236 & 9.40 & 458 \\
MonSter~\cite{cheng2025monster} & CVPR-25 & MP & \best{3.62} & \best{4.00} & \second{4.19} & \second{15.2} & 0.24 & \best{2.00} & 388.8 & 2141 & 7.64 & 595 \\
\textbf{LiteMatch} & - & - & \second{4.20} & \second{4.09} & \best{4.11} & \best{15.1} & \best{0.21} & \second{2.05} & \best{9.58} & \best{179.0} & \best{1.41} & \best{220} \\

\bottomrule
\end{tabular}}
\vspace{-3mm}
\end{table*}

\section{Experiments}
\label{sec:experiments}
\vspace{-3mm}
\subsection{Implementation Details}
\vspace{-2mm}
LiteMatch is implemented in PyTorch~\citep{paszke2019pytorch} and follows a two-stage training strategy. Training Pipeline: Our approach follows a two-stage training strategy. In Stage 1, we jointly optimize the feature extractor and cost volume modules using our proposed CVC-Loss supervision. Following Selective-IGEV's protocol, we train on Scene Flow for 1 million iterations with batch size 1 using the AdamW optimizer and a OneCycle learning rate schedule (peak $2 \times 10^{-4}$), requiring 4-5 days on a single NVIDIA RTX A6000 GPU. In Stage 2, we freeze all Stage 1 components to preserve feature stability and train exclusively the disparity refinement module using regression loss while keeping all hyperparameters consistent. This produces our base model with 3.36M parameters, while the iterative refinement variant (9.58M parameters) additionally trains the IR head during Stage 2.

\subsection{Experimental Setup}
\vspace{-2mm}
We evaluate LiteMatch on both synthetic and real-world benchmarks. All models are trained exclusively on the Scene Flow dataset~\cite{mayer2016large}, which comprises 35,454 image pairs. We compare LiteMatch against two categories of methods: stereo-only approaches (denoted as ``--'' or backbone (BB)), including RAFT-Stereo~\cite{lipson2021raft}, DLNR~\cite{zhao2023high}, IGEV-Stereo~\cite{xu2023iterative}, Selective-IGEV~\cite{wang2024selective}, and Mocha-Stereo~\cite{chen2024mocha}; and methods leveraging monocular depth priors (MP), such as DEFOM-Stereo~\cite{jiang2025defomstereo}, SMoEStereo~\cite{wang2025learning}, FoundationStereo~\cite{wen2025foundationstereo}, and MonSter~\cite{cheng2025monster}.
The zero-shot generalization performance of LiteMatch is assessed on four diverse datasets: KITTI-2012/2015~\cite{geiger2012we,menze2015object}, Middlebury-F~\cite{scharstein2014high}, ETH3D~\cite{schops2017multi}, and DrivingStereo~\cite{yang2019drivingstereo}. We report standard metrics, including End-Point Error (EPE) and D1-all ($>3$ px). For KITTI-2012, KITTI-2015, Middlebury-F, and ETH3D, we additionally report Bad X\% errors. Inference speed (FPS) is measured at a resolution of $384 \times 1248$.

\vspace{-3mm}
\begin{figure*}[!htpb]
  \centering
  \includegraphics[width=0.99\linewidth]{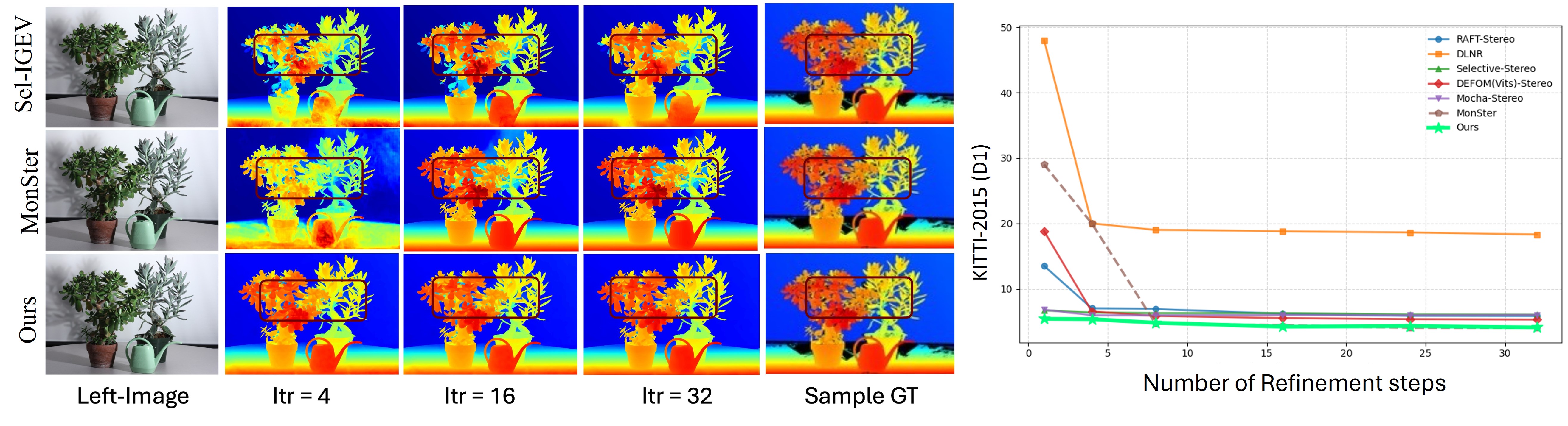}
  \vspace{-3mm}
  \caption{Iterative Refinement Comparison. Left: Disparity maps at iterations 4, 16, and 32 for MonSter, Selective-IGEV, and LiteMatch, showing convergence and structural clarity. Right: KITTI-2015 D1 error vs. refinement iterations, demonstrating LiteMatch’s early stabilization and efficient disparity refinement.
}
  \label{fig:iterative}
\end{figure*}
\begin{figure}[!ht]
  \centering
  \includegraphics[width=0.95\linewidth]{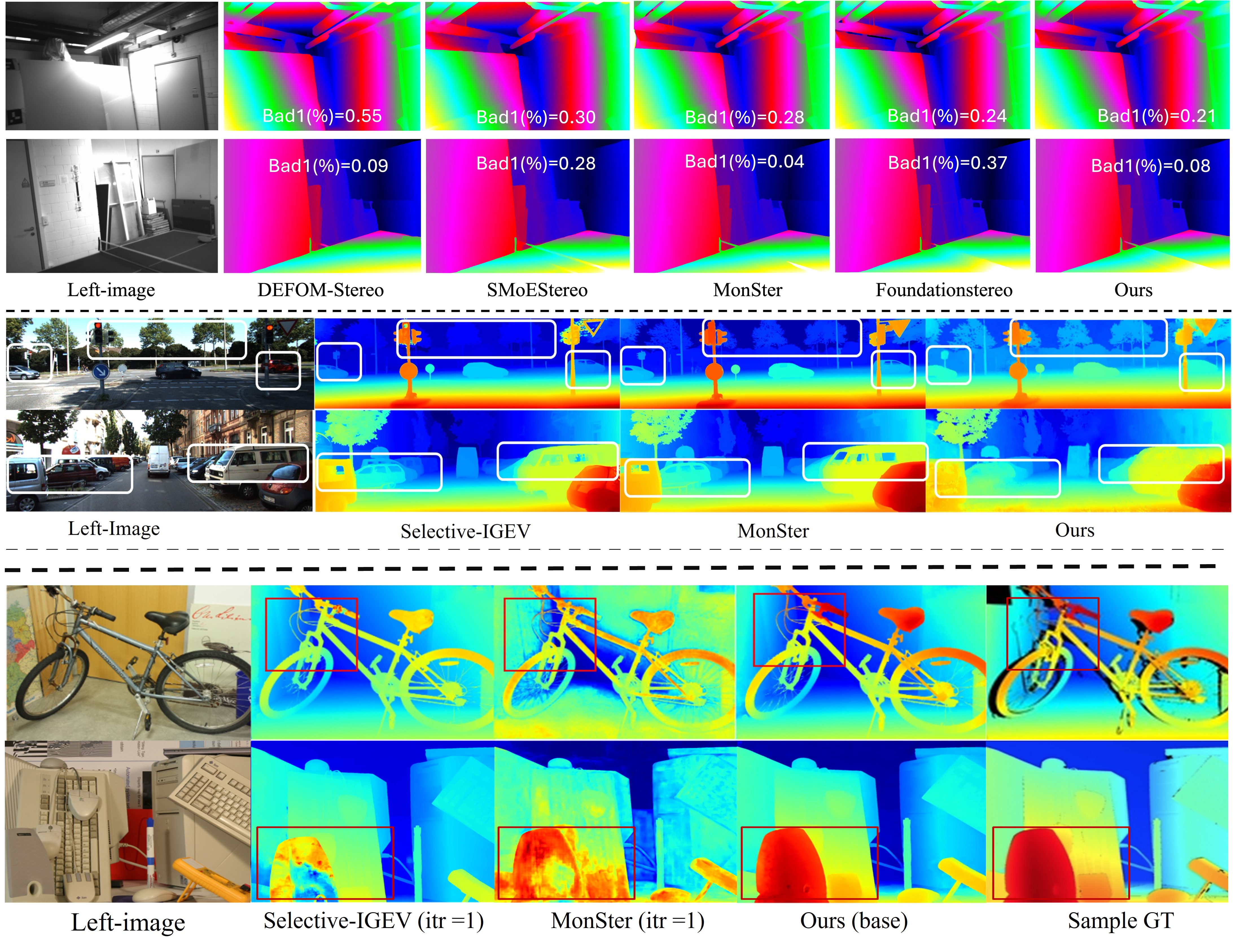}
  \caption{Qualitative comparison of LiteMatch against state-of-the-art stereo methods, illustrating sharper disparity estimation and cross-domain robustness. Rows 1–2 show results on ETH3D leaderboard samples (fine-tuned), rows 3–4 on KITTI. The final two rows present outputs from the base models and single-iteration refinements of the respective methods for reference. }
  \vspace{-5mm}
\label{fig:qualitative_comparison}
\end{figure}

\begin{table}[!ht]
  \centering
  \footnotesize
  \caption{Ablation study on model components on Scene flow datasets.}
  \vspace{-2mm}
  \label{tab:ablation-modules}
  \resizebox{\textwidth}{!}{%
  \begin{tabular}{l *{10}{c}}
    \toprule
    \multirow{3}{*}{\textbf{Model}}
    & \multicolumn{2}{c}{Feature Extraction}
    & CV
    & \multicolumn{3}{c}{Disparity Refinement}
    & \multirow{3}{*}{CVC-Loss}
    & \multirow{3}{*}{Params (M)}
    & \multirow{3}{*}{EPE $\downarrow$}
    & \multirow{3}{*}{D1 (\%) $\downarrow$} \\
    \cmidrule(lr){2-3} \cmidrule(lr){5-7}
    & HFE & CA & Refin. & Geo & Corr & IR & & & \\
    \midrule
    Baseline                  & $\times$ & $\times$ & $\times$ & \cmark & \cmark & $\times$ & $\times$ & 2.27 & 0.98 & 4.48 \\
     + CVC-Loss       & $\times$ & $\times$ & $\times$ & \cmark & \cmark & $\times$ & \cmark & 2.27 & 0.84 & 3.30 \\
    + HFE                     & \cmark   & $\times$ & $\times$ & \cmark & \cmark & $\times$ & \cmark & 2.42 & 0.70 & 3.18 \\
    + HFE + CA                & \cmark   & \cmark   & $\times$ & \cmark & \cmark & $\times$ & \cmark & 2.74 & 0.65 & 2.92 \\
    + HFE + Refin. (w/o Corr) & \cmark   & \cmark   & \cmark   & \cmark & $\times$ & $\times$ & \cmark & 3.25 & 0.59 & 2.80 \\
    \textbf{LiteMatch (Base)} & \cmark   & \cmark   & \cmark   & \cmark & \cmark & $\times$ & \cmark & \textbf{3.36} & \textbf{0.53} & \textbf{2.29} \\
    \textbf{LiteMatch}  & \cmark   & \cmark   & \cmark   & \cmark & \cmark & \cmark & \cmark & \textbf{9.56} & \textbf{0.39} & \textbf{1.92} \\
    \bottomrule
  \end{tabular}}
  \vspace{-1.5mm}
\end{table}

\begin{figure}[t]
    \centering
    \includegraphics[width=0.95\textwidth]{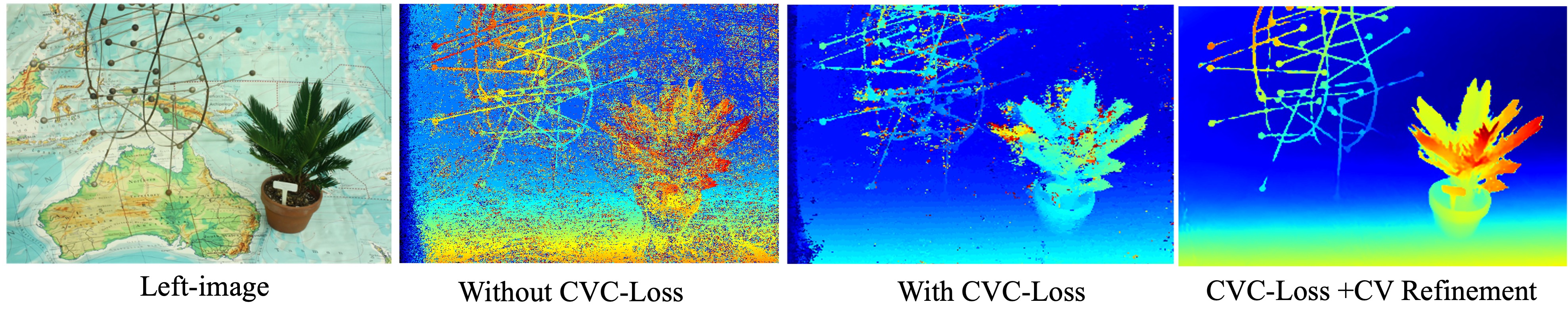}
    \vspace{-3mm}
    \caption{Visual ablation of CVC-Loss. From left to right: left input image; raw disparity without CVC-Loss (noisy, ambiguous); raw disparity with CVC-Loss (clean, sharp, low-entropy); final stage-1 refined cost volume disparity (clean and preserved details).}
    \vspace{-2mm}
    \label{fig:visual_ablation}
\end{figure} 
\begin{figure}[t]
    \centering
    \includegraphics[width=0.95\textwidth]{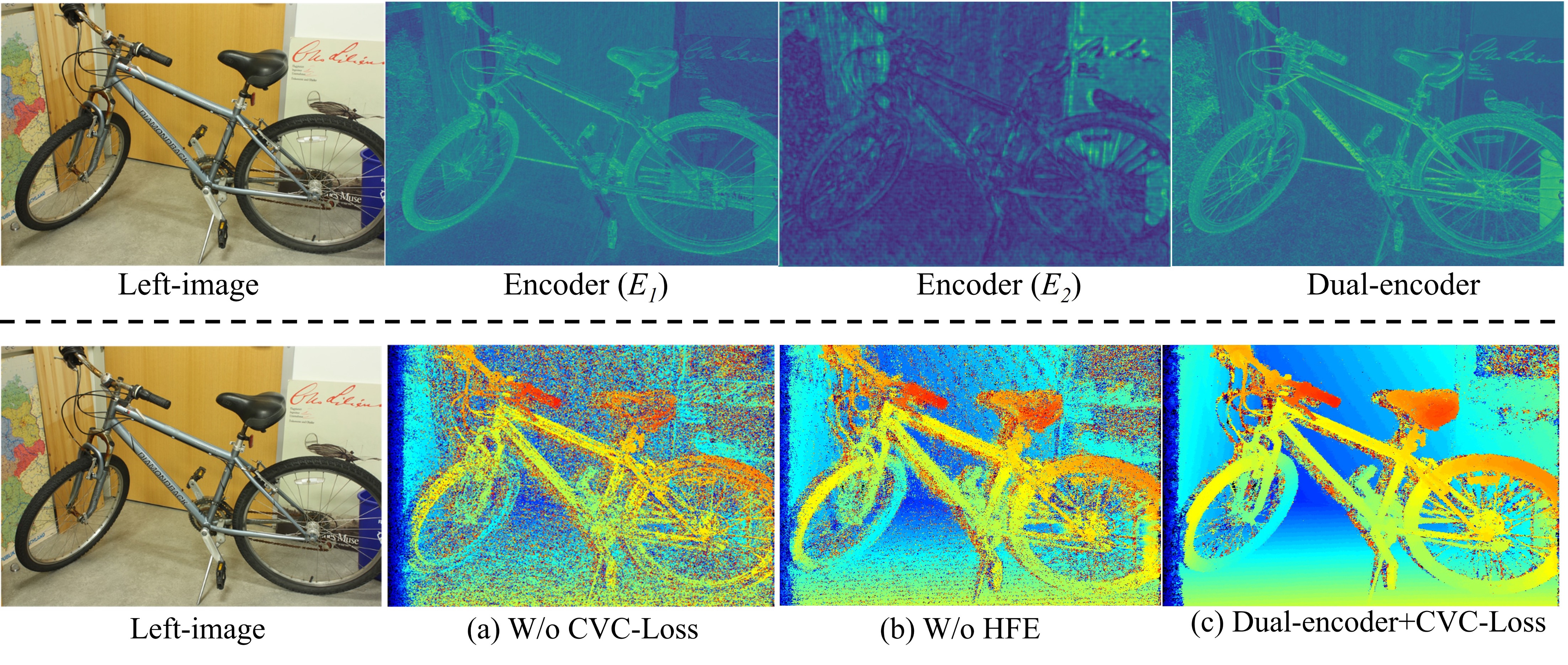}
    \vspace{-3mm}
    \caption{Top row: Complementary feature activations from the Dual Encoder($E_1$)captures smooth, globally coherent regions, while $E_2$ focuses on sparse, high-frequency edge and texture details; their fusion provides a balanced representation for disparity estimation.
Bottom row: Corresponding initial disparity maps raw correlation without CVC-Loss(a) is noisy, without HFE ($E_2$)(b) is noisy, while fused features and  CVC-Loss application  produce sharp (c), clean results, reducing the need for heavy post-processing}
    \vspace{-2mm}
    \label{fig:itr}
\end{figure}

\begin{table}[!htb]
\centering
\scriptsize
\setlength{\tabcolsep}{2pt}
\caption{
Ablation studies trained on Scene Flow.
Left: In-domain and cross-domain evaluation (HFE ablation) in terms of D1 (\%).
Right: Loss function ablation on Scene Flow.
}

\begin{minipage}{0.52\linewidth}
\centering
\begin{tabular}{lccccc}
\toprule
\multirow{2}{*}{Method} & \multirow{2}{*}{HFE} 
& \multicolumn{2}{c}{Scene Flow} 
& \multicolumn{2}{c}{DrivingStereo} \\
\cmidrule(lr){3-4} \cmidrule(lr){5-6}
 & & Edge & Non-E & Edge & Non-E \\
\midrule
LiteMatch & $\times$ & 9.45 & 3.35 & 10.62 & 3.78 \\
LiteMatch & \cmark & \textbf{7.21} & \textbf{1.42} & \textbf{8.24} & \textbf{1.85} \\
\bottomrule
\end{tabular}
\end{minipage}
\hfill
\begin{minipage}{0.44\linewidth}
\centering
\begin{tabular}{lcc}
\toprule
Stage 1 Loss & EPE $\downarrow$ &D1 (\%) $\downarrow$ \\
\midrule
Cat. CE + L1          & 0.73 & 3.25 \\
Focal ($\gamma{=}2$) + L1 & 0.64 & 2.91 \\
\textbf{CVC-Loss + L1} & \textbf{0.53} & \textbf{2.29} \\
\bottomrule
\end{tabular}
\end{minipage}

\vspace{-2mm}
\label{tab:hfe_loss_merged}
\end{table}

\noindent\textbf{Comparison with Foundation Models.}
Recent ETH3D top-performers rely on heavy hybrid designs integrating large pre-trained monocular depth foundations (MonSter 388M, FoundationStereo 375M, DEFOM-Stereo 374M) with 3D convolutions and multi-dataset training (6–8 datasets). 
In contrast, LiteMatch adopts a monocular-prior-free architecture (3.36M/9.58M) trained only on Scene Flow (35,454 pairs) with single-stage ETH3D fine-tuning. 
Despite this simplicity, LiteMatch achieves state-of-the-art ETH3D (Bad1.0: 0.42\%), surpassing FoundationStereo (0.51\%) and MonSter (0.44\%), while requiring 40$\times$ fewer parameters. 
This confirms pure stereo matching remains highly competitive with foundation-augmented approaches (see Table~S1/S2 in supplementary).

\subsection{Performance Evaluation}
\vspace{-2mm}
To validate the effectiveness of LiteMatch, we present
comprehensive results on in-domain and cross-domain (zero-shot) generalization benchmarks. We begin with in-domain performance on Scene Flow, followed by leaderboard result on ETH3D, and then evaluate zero-shot generalization across multiple real-world datasets.
In-Domain Performance on Scene Flow is shown in Table~\ref{tab:driving_sf_merged_single} where LiteMatch (9.58M) is 39$\times$ smaller than MonSter (388M) yet achieves the lowest EPE (0.39) among all methods without monocular priors and the best D1 (1.92\%) overall. It outperforms the large-scale monocular-prior models, demonstrating superior efficiency in stereo matching. Table~\ref{tab:kitti_eth3d_merged_single_best} demonstrates that LiteMatch achieves the best Bad1.0 on both ETH3D-Noc (0.35\%) and ETH3D-All (0.52 \%), with AvgErr of  0.11/0.12 and RMS of 0.19/0.31. These results position LiteMatch within the top three across all metrics, outperforming MonSter in Bad1.0 while remains competitive on other metrics, demonstrating a strong performance-complexity balance.

We further evaluate LiteMatch's robustness via zero-shot cross-domain trained on Scene Flow generalization, comparing it against both stereo-only and monocular-prior methods. As reported in Table~\ref{tab:kitti_eth3d_merged_single_best}, LiteMatch (Base) yields the best EPE (1.21/1.11) and D1 (5.40\%/5.12\%) among single-iteration stereo-only methods, coupled with the fastest inference speed of 22 FPS on KITTI-2012/2015. These results affirm LiteMatch's \textit{strong generalization to real-world driving scenarios without domain-specific fine-tuning}.
Table~\ref{tab:comparative} illustrates the results on various datasets, where LiteMatch sets a new benchmark with the lowest EPE (4.11) and Bad2.0 (15.1\%), outperforming MonSter (EPE 4.19, Bad2.0 15.2\%) on Middlebury-F despite being 39$\times$ smaller. This performance on high-resolution, dense ground-truth data illustrates LiteMatch's precision in handling complex textures and occlusions. On ETH3D challenging dataset, LiteMatch achieves the best EPE (0.21) and the second-best Bad1.0 (2.05\%), closely trailing MonSter (2.00\%). The results across varied indoor and outdoor scenes demonstrate LiteMatch's adaptability without relying on external priors.
Table~\ref{tab:driving_sf_merged_single} presents the results on the DrivingStereo dataset, where LiteMatch achieves the best D1-all (2.00\%), averaged over the entire dataset, and sets a new SOTA on the Sunny (2.01\%), Cloudy (2.05\%), and particularly the Foggy (1.51\%) subsets. This dominant performance across diverse weather and illumination conditions \textit{makes it highly suitable for real-world autonomous driving applications}.

\noindent\textbf{Convergence vs.\ Iterations:}
We evaluate the impact of iterative refinement on overall performance.
Figures~\ref{fig:iterative} illustrate the qualitative (left) and quantitative (right)  results, respectively, across different iteration counts (1, 4, \ldots, 32).
Our LiteMatch demonstrates faster convergence than existing SOTA methods, achieving strong performance even from the first iteration, whereas Selective-Stereo, DLNR, and MonSter begin to converge only after eight or more iterations.
Notably, as shown in Figure~\ref{fig:qualitative_comparison}(bottom 2 rows), our base model already outperforms Selective-Stereo and MonSter at a single iteration. \textit{This improvement may be attributed to the proposed stable feature representation and cost volume stabilization modules}.

\noindent\textbf{Performance Vs Complexity Analysis:}
\label{sec:tradeoff}
From Table~\ref{tab:comparative}, our model achieves an optimal trade-off between performance and complexity over SOTA methods. Without monocular depth priors, our LiteMatch (Base) model (3.36M params, 79.8~GFLOPs, computed including the FFT filtering module) surpasses all SOTA in inference speed (22.2~FPS vs.\ 2.8--16.1~FPS for baselines) while maintaining SOTA performance (5.12\% Bad3.0 on KITTI-2012, better than IGEV-Stereo's 6.67\%).
When extended with iterative refinement (LiteMatch), it further closes the performance gap to monocular-prior-based methods (4.09\% Bad3.0 vs.\ MonSter's 4.00\%) but remains 2.5$\times$ faster (220~ms vs.\ 595~ms) and 39$\times$ lighter (9.58M vs.\ 388.8M params).
This efficiency combined with minimal memory footprint (1.17--1.41~GB vs\ 7.08--9.4~GB in SOTA) makes our approach \textit{uniquely suited for real-time applications on resource-constrained platforms, from autonomous systems to edge devices}.

\vspace{-3mm}
\subsection{Ablation Study}
\vspace{-2mm}
We perform ablation studies on Scene Flow to evaluate each component of LiteMatch, reporting EPE and D1-all (\%). As shown in Table~\ref{tab:ablation-modules}, the baseline using only the Cross-view Correspondence Encoder (E1) and non-iterative refinement achieves 0.98 EPE / 4.48\% D1, indicating unstable raw correlation. Adding CVC-Loss reduces EPE to 0.84 by stabilizing disparity distributions through dense negative supervision. Incorporating the High-Frequency Encoder (HFE) further improves performance to 0.70 EPE by enhancing structural fidelity at edges and thin regions. Introducing cross-attention (CA) lowers EPE to 0.65, confirming the complementarity of global correspondence modeling and frequency-aware features. With cost volume refinement, EPE decreases to 0.59, while adding the Correspondence Encoder completes LiteMatch (Base) at 0.53 EPE / 2.29\% D1. Enabling the optional iterative refinement (IR) head achieves the best performance of 0.39 EPE / 1.92\% D1, demonstrating that stabilized representations support efficient multi-step refinement.

Fig.~\ref{fig:visual_ablation} compares raw and refined disparities with and without CVC-Loss. 
Without CVC-Loss, the raw disparity is noisy and ambiguous. 
With CVC-Loss, it becomes clean and sharp, while refinement recovers high-frequency details, confirming that early-stage stabilization reduces the burden on refinement. Figure~\ref{fig:itr} qualitatively supports these results. E1 captures smooth global structures, whereas HFE emphasizes high-frequency edge details; their fusion yields balanced representations. Without CVC-Loss (Fig.~\ref{fig:itr}(a)), disparity initialization is noisy; without HFE (Fig.~\ref{fig:itr}(b)), boundaries are blurred. The full model (Fig.~\ref{fig:itr}(c)) produces sharp and well-localized disparities, indicating that feature complementarity and cost stabilization reduce ambiguity at initialization.

Table~\ref{tab:hfe_loss_merged} (left) shows that HFE improves edge-region accuracy both in-domain and cross-domain (Scene Flow: 9.45\% $\rightarrow$ 7.21\%; DrivingStereo: 10.62\% $\rightarrow$ 8.24\%), highlighting improved structural generalization. Table~\ref{tab:hfe_loss_merged} (right) compares Stage~1 supervision strategies: CVC-Loss + L1 outperforms CE and Focal Loss, reducing EPE from 0.64 to 0.53 and D1 from 2.91\% to 2.29\%. Figure~\ref{fig:entropy_distribution} further shows that CVC-Loss produces lower-entropy and more concentrated disparity distributions, confirming its stabilizing effect.
\vspace{-4mm}
\section{Conclusion}
\vspace{-5mm}
This paper presented LiteMatch, a lightweight stereo matching framework that achieves SOTA performance without relying on monocular depth priors or large pre-trained backbones. Our approach addresses the fundamental limitation of cost volume instability through two key contributions: a novel CVC-Loss that stabilizes matching confidence during early training, and a dual-encoder architecture that extracts globally coherent yet spatially precise features. For disparity refinement, we proposed NIR module that refines with single forward pass with an optional IR head serves
as a lightweight alternative to SOTA iterative methods. Extensive experiments demonstrate that LiteMatch establishes aan improved balance between performance and complexity. With only 9.58M parameters (39 × fewer than MonSter) our method achieves superior performance on Scene Flow (1.92\% D1) and exceptional zero-shot generalization across multiple benchmarks including KITTI, DrivingStereo, Middlebury, and ETH3D. LiteMatch proves that principled architectural design and targeted stabilization strategies can eliminate the dependency on computationally expensive components, enabling high-performance stereo matching for resource-constrained applications.

\noindent\textbf{Acknowledgement:}\\
This research was supported by 
(d-real) Science Foundation Ireland (Grant 18/CRT/6224).
The authors thank the CVPR Lab members at Trinity Col-
lege Dublin and IIT Ropar for their support.
\vspace{-3mm}
\clearpage
\input{supp}

\bibliographystyle{splncs04}
\bibliography{main}

\end{document}

%% file: supp.tex
\title{LiteMatch: Lightweight Zero-Shot Stereo Matching via Cost Volume Stabilization \\ \vspace{0.5em} Supplementary Material}

\titlerunning{LiteMatch Supplementary Material}

\author{Md Raqib Khan$^{1}$ \and
        Santosh Kumar Vipparthi$^{2}$ \and
        Subrahmanyam Murala$^{1}$}

\institute{CVPR Lab, Trinity College Dublin, The University of Dublin, Dublin, Ireland \and 
CVPR Lab, Indian Institute of Technology Ropar, Rupnagar, Punjab, India\\
\email{khanmd@tcd.ie}}

\authorrunning{Khan et al.}

\maketitle

\section*{Overview}
This supplementary document provides additional comparison with SOTA, convergence analyses, and additional implementation details for the LiteMatch framework. It includes:
\begin{itemize}
    \item Section~\ref{sec:sota}: Additional Comparison with SOTA
    \item Section~\ref{sec:analysis}: Convergence Analysis
    \item Section~\ref{sec:Implementation}: Additional Implementation Details
\end{itemize}

\setcounter{section}{0}
\renewcommand{\thesection}{S\arabic{section}}
\renewcommand{\thefigure}{S\arabic{figure}}
\renewcommand{\thetable}{S\arabic{table}}

\section{Additional Comparison with SOTA} \label{sec:sota}
\begin{table*}[t]
\centering
\tiny
\setlength{\tabcolsep}{5pt}
\caption{Comparison of leading stereo matching architectures based on their core components, highlighting the distinctions between existing methods and ours}
\begin{tabular}{lccccc}
\toprule
\textbf{Method} & \textbf{Backbone} & \textbf{3D Conv} & \textbf{Iterative Refinement}& \textbf{$\#$param(M)} \\
\midrule
RAFT-Stereo (3DV'21) & --& No & Yes&11.1 \\
IGEV-Stereo (CVPR'23) &  MobileNetV2-100& Yes& Yes&12.6 \\
IGEV++ (TPAMI'2024) &  MobileNetV2-100 & Yes & Yes&12.4\\
Selective-IGEV (CVPR'2024) & MobileNetV2-100 & Yes& Yes &13.1\\
DEFoM-Stereo (CVPR'25) &Mono Depth Pre-trained& Yes & Yes &374 \\
SMoE-Stereo (ICCV'2025) & Mono Depth Pre-trained & Yes & Yes&113 \\
MonSter (CVPR'2025) & Mono Depth Pre-trained  & Yes& Yes &388\\
FoundationStereo (CVPR'25) & Mono Depth Pre-trained & Yes & Yes&375 \\

\textbf{LiteMatch (base)} & -- & No & No&\textbf{3.36} \\
\textbf{LiteMatch} & -- & No & Yes&\textbf{9.58} \\
\bottomrule
\label{tab:comparison_architecure}
\end{tabular}
\end{table*}

Most recent top-performing methods on ETH3D rely on heavy hybrid designs that integrate large pre-trained monocular depth foundation models (DepthAnythingV2 in DEFOM-Stereo, SMoE-Stereo, MonSter, and FoundationStereo) for feature extraction or initialization, combined with 3D-convolution cost-volume processing and iterative refinement, resulting in parameter counts of 113M--388M (Table~\ref{tab:comparison_architecure}). These approaches further exploit extensive multi-dataset pre-training and fine-tuning schedules involving 6--8 mixed synthetic/real datasets  including{(TartanAir~\cite{tartanair2020iros}, CREStereo~\cite{li2022practical}, InStereo2k~\cite{bao2020instereo2k}, and ETH3D itself)} or even custom {1M-pairs} (Table~\ref{tab:dataset_consiseration}), effectively injecting strong monocular priors and domain-specific knowledge that go far beyond pure stereo matching. In contrast, LiteMatch adopts a lightweight, monocular-prior-free architecture (3.36M parameters base, 9.58M with lightweight iterative refinement) and follows a minimal training protocol: pre-training only on the standard Scene Flow dataset (35,454 pairs) followed by single-stage fine-tuning exclusively on ETH3D. Despite this extreme simplicity and the absence of any external monocular or multi-dataset augmentation, LiteMatch achieves state-of-the-art ETH3D, surpassing hybrid foundations such as FoundationStereo (375M parameters, {1M synthetic pairs + DepthAnythingV2 priors)}, demonstrating that pure, data-efficient stereo matching remains highly competitive with contemporary heavily augmented baselines.

\begin{table*}[!h]
\centering
\caption{Comparison of training strategies for stereo matching. Our method employs a straightforward pipeline, pre-training on a single synthetic dataset (Scene Flow) and fine-tuning only on the target benchmark (ETH3D). This contrasts with the prevalent trend of using large-scale multi-dataset mixtures for pre-training and/or fine-tuning.}
\label{tab:dataset_consiseration}
\small
\setlength{\tabcolsep}{1pt}
\begin{tabular}{lccc}
\toprule
\textbf{Method} & \textbf{Pre-training} & \textbf{Finetuning Strategy} & \textbf{Finetuning Datasets}  \\
\midrule
RAFT-Stereo~\cite{lipson2021raft} & Scene Flow & Single-stage & ETH3D \\
Selective-IGEV~\cite{wang2024selective} & Scene Flow & Multi-stage & 6 datasets  \\
DEFOM-Stereo~\cite{jiang2025defomstereo} & Scene Flow & Two-stage &  6 datasets  \\
SMoE-Stereo~\cite{wang2025learning} & Scene Flow & Single-stage & 4 real datasets  \\
MonSter~\cite{cheng2025monster} & 8 datasets (mixed) & Multi-stage & Heavy mixture  \\
FoundationStereo~\cite{wen2025foundationstereo} & FS80K ($\sim$1M pairs) & Multi-stage & Multiple datasets  \\
\midrule
\textbf{Ours} & \textbf{Scene Flow} & \textbf{Single-stage} & \textbf{ETH3D only} \\
\bottomrule
\end{tabular}
\end{table*}

\section{Convergence Analysis} \label{sec:analysis}
\begin{table*}[!ht]
\tiny
\setlength{\tabcolsep}{2pt}
\caption{Bad 3.0 (\%) across refinement iterations on KITTI-2015 for models trained on Scene Flow. Our method reaches convergence in fewer iterations.}
 
\begin{tabular}{c|cccccccc}
\toprule

\textbf{Iter} & \textbf{RAFT-Stereo} & \textbf{IGEV} & \textbf{DLNR} & \textbf{Selective-IGEV} & \textbf{DEFoM-Stereo} & \textbf{Mocha-Stereo} & \textbf{MonSter}& \textbf{Ours} \\
\midrule
1  & 13.50 & 6.64 & 48.0 & 6.68 & 18.8 & 6.80 & 29.0 & 5.35 \\
4  & 7.01  & 6.40 & 20.0 & 6.38 & 6.54 & 5.95 & 20.0 & 4.50 \\
8  & 6.91  & 6.28 & 19.0 & 6.27 & 5.84 & 5.94 & 4.70 & 4.30 \\
16 & 6.17  & 6.15 & 18.8 & 6.27 & 5.51 & 5.99 & 4.40 & 4.20 \\
24 & 5.86  & 6.04 & 18.6 & 6.09 & 5.35 & 5.95 & 4.03 & 4.12 \\
32 & 5.83  & 6.03 & 18.3 & 6.06 & 5.29 & 6.01 & 4.00 & 4.09 \\
\midrule
Converges at. & 32 & 32 & 32 & 16 & 32 & 8 & 32 & \textbf{4} \\
\bottomrule
\end{tabular}
\label{tab:convergence}
\end{table*}
Table~\ref{tab:convergence} reports the Bad 3.0 (\%) across refinement
iterations for models trained on Scene Flow and evaluated on KITTI-2015.
To quantify convergence speed, we define convergence as the smallest iteration
$n$ for which the relative change in EPE between two successive iterations is
less than 1\%:

\begin{equation}
\frac{|EPE_n - EPE_{n-1}|}{EPE_{n-1}} < 1\%.
\end{equation}

Under this criterion, our method converges after only four iterations, whereas
existing approaches typically require between 8 and 32 iterations. This
demonstrates that our refinement process reaches a stable solution significantly
faster while maintaining strong performance when generalizing from synthetic
(Scene Flow) training to real-world KITTI-2015 datasets.

\section{Additional Implementation Details}
\label{sec:Implementation}
\vspace{-2mm}
\noindent\textbf{Additional training details:}
For ETH3D fine-tuning, we adapt our training configuration to the benchmark's high-resolution characteristics. Input images are processed at resolution $384 \times 768$. We employ a learning rate of $1\times10^{-4}$ upto 40k iterations then reduced to $5\times10^{-5}$ upto 100k , maintaining a batch size of 1, This conservative schedule ensures stable convergence while adapting our SceneFlow-pretrained features.

\noindent\textbf{Additional details for training strategy:}
We employ two stage training to address optimization challenges in end-to-end stereo learning.

\noindent\textbf{Why the two-stage (Stage-1 $\to$ Stage-2) schedule works?}
\label{supp:training-rationale}Let $\theta$ denote Stage-1 parameters (dual-encoder, cost-volume refinement) producing
features $F_L,F_R$, cost volumes $C_0,C_2$, and coarse disparities $D_0,D_2$.
Let $\phi$ denote Stage-2 parameters (E\textsubscript{geo}, E\textsubscript{corr}, decoder, IR).
The full joint objective optimized in principle is
\begin{equation}
    \min_{\theta,\phi}\; L(\theta,\phi)
    \;=\; L_{\text{stage1}}(\theta) \;+\; L_{\text{stage2}}(\theta,\phi),
\end{equation}
where (following the paper) $L_{\text{stage1}}$ contains the Smooth-$L_1$ terms on $D_0,D_2$ and the CVC-loss on cost volumes (Sec.~3.2), and $L_{\text{stage2}}$ contains the regression losses for the refinement outputs $D_{\text{base}}, D_1,\dots,D_n$.

\noindent\textbf{Non-stationarity and gradient-variance.}
The refinement receives input
\[
g(x;\theta)=\big(C_2(\theta),\, f_L^4(\theta),\, D_2(\theta)\big).
\]
During naive joint training both $\theta$ and $\phi$ are updated simultaneously. Updates to $\theta$ change the distribution of $g(x;\theta)$ while $\phi$ is learning, inducing \emph{non-stationary} training inputs for $\phi$~\cite{bottou2018optimization}. For SGD, the variance of the stochastic gradient for $\phi$ scales with the variability of $g$:
\[
\mathrm{Var}\big[\nabla_\phi L_{\text{stage2}}(g(x;\theta),\phi)\big]
\propto \mathrm{Var}\big[g(x;\theta)\big]~\text{\cite{faghri2020study}}.
\]
High variance slows convergence, increases required damping (smaller LR), and can cause oscillations or collapse of learned refinements~\cite{li2022practical}.

\noindent\textbf{Two-stage decomposition (algorithmic benefits).}
The practical two-stage schedule:
\[
\begin{aligned}
&\theta^\star\;\approx\;\arg\min_\theta L_{\text{stage1}}(\theta),\\
&\phi^\star\;\approx\;\arg\min_\phi L_{\text{stage2}}(\theta^\star,\phi),\\
&\text{(optional)}\quad (\theta,\phi)\leftarrow \text{few low-LR steps from }(\theta^\star,\phi^\star),
\end{aligned}
\]
Our two-stage training strategy yields several important advantages. First, freezing $\theta^\star$ after Stage-1 removes input drift, ensuring that $\phi$ observes a fixed feature extractor $g(\cdot;\theta^\star)$ throughout refinement. This reduces gradient variance and enables the refinement network to learn consistent residual corrections~\cite{pang2017cascade}. Second, solving $L_{\text{stage1}}$ first biases $\theta$ into an optimization basin characterized by sharp, low-entropy cost volumes a property explicitly enforced by our CVC-loss. Initializing the refinement parameters $\phi$ within this favorable regime positions the joint parameter set $(\theta^\star,\phi^\star)$ for more effective final finetuning~\cite{weinshall2018curriculumlearningtransferlearning}. Third, freezing the backbone prevents destructive gradient updates to the geometric prior encoded by $\theta$, allowing $\phi$ to learn stable correction operators rather than continually adapting to a moving target~\cite{li2022practical}. Finally, this staged approach embodies a natural coarse-to-fine curriculum: Stage-1 establishes globally coherent matching structure, while Stage-2 refines local details. This decomposition simplifies each learning task and empirically improves both convergence speed and final accuracy~\cite{weinshall2018curriculumlearningtransferlearning}.

\noindent\textbf{Analytic sketch (variance reduction).}
Let $G_\phi(x;\theta,\phi)=\nabla_\phi \ell\big(g(x;\theta),\phi\big)$ be the per-sample gradient. If $\theta$ varies across SGD steps, the variance of $G_\phi$ increases:
\[
\mathrm{Var}[G_\phi] = \mathbb{E}_x\!\big[\|G_\phi\|^2\big] - \|\mathbb{E}_x[G_\phi]\|^2,
\]
and the first term grows when $g(x;\theta)$ distribution shifts~\cite{bottou2018optimization}. Fixing $\theta$ eliminates this source of shift, lowering $\mathrm{Var}[G_\phi]$, and thereby improving convergence speed and stability when training $\phi$~\cite{faghri2020study}.

\noindent\textbf{Training strategy for LiteMatch.}
Motivated by the success of cascaded architectures, we decompose training into two distinct phases to stabilize learning and specialize module functions. \textbf{Stage-1} focuses on learning robust feature correspondence and an initial disparity estimate by optimizing the combined objective $L_{\text{stage1}}$ on SceneFlow. \textbf{Stage-2} then specializes the refinement network; with the Stage-1 parameters $\theta^{\star}$ frozen, it is trained solely with $L_{\text{stage2}}$ to enhance the resolution and precision of the high-resolution output, preventing the refinement process from interfering with the already-learned robust matching capabilities.